\documentclass[10pt,twocolumn,letterpaper]{article}

\usepackage[pagenumbers]{cvpr}  % Enable page numbers for arXiv and camera-ready

\usepackage{times}
\usepackage{epsfig}
\usepackage{graphicx}
\usepackage{amsmath}
\usepackage{amssymb}
\usepackage{booktabs}
\usepackage{multirow}
\usepackage{microtype}
\usepackage{nicefrac}
\usepackage{amsfonts}
\usepackage{url}
\usepackage{natbib}

\usepackage{xcolor}
\definecolor{cvprblue}{rgb}{0.21,0.49,0.74}
\usepackage[pagebackref,breaklinks,colorlinks,allcolors=cvprblue]{hyperref}

\def\confYear{2025}

\title{FedMM-X: A Trustworthy and Interpretable Framework for Federated Multi-Modal Intelligence in Dynamic Environments}

\author{
Sree Bhargavi Balija \\
University of California, San Diego \\
Department of Electrical and Computer Engineering \\
{\tt\small sbalija@ucsd.edu}
}

\begin{document}

\maketitle

%%%%%%%%% ABSTRACT
\begin{abstract}
As artificial intelligence systems increasingly operate in Real-world environments, the integration of multi-modal data sources such as vision, language, and audio presents both unprecedented opportunities and critical challenges for achieving trustworthy intelligence. In this paper, we propose a novel framework that unifies federated learning with explainable multi-modal reasoning to ensure trustworthiness in decentralized, dynamic settings. Our approach, called FedMM-X (Federated Multi-Modal Explainable Intelligence), leverages cross-modal consistency checks, client-level interpretability mechanisms, and dynamic trust calibration to address challenges posed by data heterogeneity, modality imbalance, and out-of-distribution generalization. Through rigorous evaluation across federated multi-modal benchmarks involving vision-language tasks, we demonstrate improved performance in both accuracy and interpretability while reducing vulnerabilities to adversarial and spurious correlations. Further, we introduce a novel trust score aggregation method to quantify global model reliability under dynamic client participation. Our findings pave the way toward developing robust, interpretable, and socially responsible AI systems in Real-world environments.
\end{abstract}
\section{Introduction}

The proliferation of multi-modal artificial intelligence (AI) systems those capable of jointly understanding visual, linguistic, auditory, and other sensory inputs has unlocked new frontiers in Real-world applications, including autonomous navigation, healthcare diagnostics, and interactive virtual assistants. However, as these systems scale to real-world deployments, a critical issue remains largely unresolved: \textit{how can we ensure their trustworthiness} when trained on decentralized, incomplete, and potentially non-i.i.d. data streams

Trust in AI encompasses multiple dimensions: \textbf{robustness}, \textbf{interpretability}, \textbf{fairness}, and \textbf{generalization} to novel, unseen scenarios. In Real-world settings, where new data modalities, user contexts, and environmental conditions continuously evolve, these challenges are amplified. Moreover, centralizing such sensitive multi-modal data for training presents serious concerns related to privacy, scalability, and bias propagation. These issues call for a paradigm shift one that jointly supports \textit{federated learning (FL)} \cite{a1,a2,a3,a4,a5} for privacy-preserving model training, \textit{explainable AI (XAI)} for human-aligned transparency, and \textit{multi-modal reasoning} for contextual understanding.

Recent efforts in federated learning have shown promising results for image and text modalities separately, but most models assume static environments and unimodal inputs. On the other hand, multi-modal transformers and vision-language pretraining techniques achieve state-of-the-art performance in benchmark tasks but often lack interpretability and robustness under distribution shifts or client-specific biases. Furthermore, the lack of \textit{semantic alignment} across modalities and clients makes it difficult to build models that are both \textbf{accountable} and \textbf{trustworthy}.

To bridge this gap, we propose \textbf{FedMM-X (Federated Multi-Modal Explainable Intelligence)}, a novel framework that integrates interpretable reasoning into a federated multi-modal learning setup. Our framework addresses three core challenges in navigating trust in Real-world multi-modal AI:

\begin{itemize}
    \item \textbf{Modality-aware Personalization}: Clients differ in terms of available data modalities (e.g., some may only have vision or text), leading to incomplete modality overlap. FedMM-X introduces a cross-modal distillation mechanism to ensure mutual knowledge sharing and robust learning, even in partial-modality settings.
    \item \textbf{Interpretability at the Edge}: By deploying lightweight \textit{Neural Additive Models (NAMs)} and attention attribution techniques on each client, we enable human-understandable explanations per modality, allowing domain experts to understand what drives local decisions.
    \item \textbf{Global Trust Aggregation}: We propose a novel \textit{trust score calibration module}, which aggregates both model confidence and explanation consistency across clients to inform global model updates, making the central model more reliable in dynamic participation scenarios.
\end{itemize}

Our approach is benchmarked on multiple federated multi-modal datasets, including vision-language tasks like federated VQA (Visual Question Answering) and image captioning under client drift. Results show that FedMM-X significantly improves interpretability metrics without compromising performance, while also demonstrating increased resilience to adversarial noise and missing modalities.

\noindent In summary, we argue that ensuring \textbf{trustworthiness in multi-modal Real-world intelligence} requires a fundamental rethinking of how models are trained, interpreted, and evaluated across distributed, heterogeneous, and privacy-sensitive environments. FedMM-X offers a practical and scalable step in this direction, setting the stage for next-generation trustworthy AI systems.

\section{Background and Related Work}

\textbf{Federated Learning (FL)} has emerged as a promising paradigm for privacy-preserving distributed machine learning by enabling model training across decentralized data sources without requiring raw data transfer \cite{mcmahan2017communication}. While FL has been successfully applied in image and text domains, its extension to multi-modal settings remains relatively underexplored, especially in environments where clients possess heterogeneous modality subsets. Classical FL algorithms such as FedAvg \cite{mcmahan2017communication} and FedProx \cite{li2020federated} do not account for modality imbalance or semantic misalignment across clients.

\noindent\textbf{Multi-Modal Learning (MML)} focuses on jointly modeling heterogeneous data such as vision and language. Transformer-based architectures like CLIP \cite{radford2021learning}, Flamingo \cite{alayrac2022flamingo}, and VisualBERT \cite{li2019visualbert} have shown impressive performance in multi-modal understanding tasks. However, these models rely on large-scale centralized data and lack transparency in decision-making, which is crucial in real-world Real settings. They also assume static modality configurations, making them unsuitable for federated or dynamic participation scenarios.

\noindent\textbf{Explainable AI (XAI)} has gained traction for increasing trust and transparency in AI models. Methods like LIME \cite{ribeiro2016lime}, SHAP \cite{lundberg2017shap}, and attention visualization \cite{chefer2021transformer} provide insights into model behavior, but integrating such techniques into federated multi-modal pipelines is challenging due to communication costs and lack of standardized interpretability metrics across clients and modalities.

\noindent Recent work has explored \textbf{Federated Multi-Modal Learning}, such as FedPercept \cite{wang2022fedpercept} and FedFusion \cite{zhu2021federated}, which attempt to learn joint representations from clients with different modalities. However, these methods still treat interpretability as an afterthought and struggle under modality drift or adversarial settings.

Our proposed framework, FedMM-X, is motivated by the need to unify these lines of research federated learning, multi-modal modeling, and explainability into a cohesive system that ensures trustworthiness, robustness, and interpretability in Real-world AI environments.

\section{Motivation and Challenges}

While significant progress has been made in federated learning, multi-modal representation learning, and explainable AI, their integration into a single, trustworthy framework for dynamic, real-world environments remains limited. Most existing approaches assume fixed modality availability, centralized training data, or homogeneous client behavior conditions that rarely hold in practical deployments.

\noindent\textbf{Challenge 1: Modality Heterogeneity.}  
In many applications, clients may have access to only a subset of modalities due to hardware constraints or data availability (e.g., images on mobile devices, text data in chatbots). Existing FL methods often assume uniform input modalities, which reduces generalization capability in real-world scenarios.

\noindent\textbf{Challenge 2: Limited Explainability in Federated Multi-Modal Models.}  
State-of-the-art multi-modal models (e.g., CLIP, Flamingo) provide little to no insight into why a prediction was made, especially when deployed in federated settings. This lack of transparency hinders trust and adoption in critical domains like healthcare, finance, and autonomous systems.

\noindent 
These challenges motivate the design of \textbf{FedMM-X}, a unified framework that (i) supports modality-incomplete federated clients, (ii) integrates interpretable reasoning at the edge, and (iii) introduces trust-aware aggregation mechanisms to improve robustness under uncertainty and adversarial participation.

\section{Problem Formulation}

We consider a federated multi-modal learning setup composed of a set of $K$ clients, denoted as $\mathcal{C} = \{C_1, C_2, \ldots, C_K\}$, each holding private data sampled from potentially non-i.i.d. and modality-incomplete distributions. Let $\mathcal{D}_k = \{(x_i^{(k)}, y_i^{(k)})\}_{i=1}^{n_k}$ represent the local dataset on client $C_k$, where $x_i^{(k)}$ is a multi-modal input vector (e.g., image, text, audio) and $y_i^{(k)}$ is the corresponding label. Importantly, the set of modalities available on client $C_k$, denoted $\mathcal{M}_k$, may differ from those on other clients, i.e., $\mathcal{M}_k \neq \mathcal{M}_{k'}$ for some $k \neq k'$.
\noindent
The central server coordinates global learning by periodically aggregating model parameters or updates from participating clients. However, several challenges arise:

\begin{enumerate}
    \item \textbf{Modality Incompleteness}: Clients may possess only a subset of modalities (e.g., vision-only or text-only), complicating global fusion and consistency across heterogeneous views.
    \item \textbf{Non-IID and Real-World Data}: Client data distributions may shift over time or contain classes/modalities not seen during pretraining, affecting generalization and stability.
    \item \textbf{Uncertainty and Trust}: Aggregating predictions from unreliable or adversarial clients may degrade the global model. There is a need to quantify both local explanation reliability and global trust.
\end{enumerate}

To address these challenges, we define the objective of \textbf{Trustworthy Federated Multi-Modal Learning} as the joint optimization of:

\begin{itemize}
    \item \textbf{Predictive Performance:} Minimize empirical loss $\mathcal{L}_{\text{pred}} = \sum_{k=1}^{K} \mathbb{E}_{(x, y) \sim \mathcal{D}_k}[\ell(f_k(x), y)]$ where $f_k$ is the local model on client $C_k$ and $\ell$ is the prediction loss function (e.g., cross-entropy).
    \item \textbf{Cross-Modal Consistency:} Ensure that model decisions remain consistent across available modalities on each client, measured via a consistency regularization term $\mathcal{L}_{\text{modal}}$.
    \item \textbf{Local Interpretability:} Provide feature-attribution explanations $E_k(x)$ for each input $x$ on client $C_k$ using models such as Neural Additive Models (NAMs) or attention attribution.
    \item \textbf{Trust-Aware Aggregation:} Assign dynamic trust scores $T_k$ to clients based on explanation consistency, confidence calibration, and historical performance, and use these to weight global aggregation.
\end{itemize}

The overall goal of our framework, FedMM-X, is to collaboratively train a global multi-modal model $f_G$ that achieves:
\[
\min_{f_G} \sum_{k=1}^{K} T_k \left[ \mathcal{L}_{\text{pred}}^{(k)} + \lambda_1 \mathcal{L}_{\text{modal}}^{(k)} + \lambda_2 \mathcal{L}_{\text{intp}}^{(k)} \right]
\]
where $\lambda_1, \lambda_2$ are tunable weights for consistency and interpretability regularization.

This formulation ensures that the resulting model is not only performant and privacy-aware but also interpretable, trustworthy, and robust to real-world variability in client participation and modality availability.

\section{Datasets}

To evaluate the performance, interpretability, and robustness of our proposed \textbf{FedMM-X} framework, we utilize two benchmark multi-modal datasets adapted to a federated setting:

\noindent\textbf{1. VQA-Fed (Federated Visual Question Answering):}  
We construct a federated version of the VQA v2 dataset, where each client receives a unique subset of image-question-answer triplets. To simulate modality heterogeneity, we randomly assign clients access to either image-only, text-only, or both modalities. Each client trains on its local subset, and the global model is evaluated on the complete test set. This setup enables us to test the system's ability to learn joint representations and maintain reasoning consistency across modalities.

\noindent\textbf{2. Fed-MSCoco (Federated MSCOCO Captioning):}  
We extend the MSCOCO 2017 dataset  to a federated image captioning scenario. Clients are partitioned based on object categories and caption style (e.g., factual vs. fdescriptive), introducing non-i.i.d. conditions. Some clients only have access to image data, while others only to captions. Our evaluation focuses on caption quality, modality-aligned reasoning, and explanation faithfulness.

\noindent\textbf{3. Fed-AVQA (Audio-Visual Question Answering):}  
To test the generality of FedMM-X in truly multi-modal setups, we use the Audio-Visual VQA dataset, where each client receives examples with varying combinations of audio, video, and text. We simulate missing modalities and apply interpretability mechanisms per input stream, assessing both prediction accuracy and explanation coherence.

Across all datasets, we simulate federated learning with 10 to 20 clients, enabling experiments on trust-aware aggregation, interpretability at the edge, and robustness to adversarial client behaviors.
\section{Experimentation and Results}

We evaluate our proposed \textbf{FedMM-X} framework across three key dimensions: \textbf{predictive performance}, \textbf{interpretability}, and \textbf{trust-aware robustness}. Experiments are conducted on the VQA-Fed, Fed-MSCoco, and Fed-AVQA datasets with 10–20 clients, simulated partial modality availability, and non-i.i.d. data distributions.

\subsection{Predictive Performance}

We measure task-specific accuracy for classification-based VQA and BLEU/CIDEr scores for image captioning tasks. Table~\ref{tab:performance} summarizes the performance of FedMM-X compared to baseline methods including FedAvg, FedPercept \cite{wang2022fedpercept}, and centralized CLIP-based models. FedMM-X consistently outperforms other federated baselines, achieving a 3--5\% improvement in accuracy and caption quality across all datasets.

\begin{table}[ht]
\centering
\caption{Performance comparison of FedMM-X with baselines.}
\label{tab:performance}
\begin{tabular}{lccc}
\hline
\textbf{Method} & \textbf{VQA Acc (\%)} & \textbf{BLEU-4} & \textbf{CIDEr} \\
\hline
FedAvg & 59.3 & 0.292 & 0.89 \\
FedPercept & 61.1 & 0.310 & 0.95 \\
\textbf{FedMM-X (Ours)} & \textbf{64.7} & \textbf{0.336} & \textbf{1.12} \\
\hline
\end{tabular}
\end{table}

\subsection{Interpretability Evaluation}

To evaluate interpretability, we apply attention attribution and Neural Additive Models (NAMs) on each client. We compute \textbf{Explanation Consistency (EC)} as the cosine similarity between explanation vectors of different modalities for the same input, and \textbf{Faithfulness Score (FS)} using perturbation tests. FedMM-X shows a 12\% improvement in EC and a 9.5\% increase in FS over FedPercept.

\begin{table}[ht]
\centering
\caption{Interpretability metrics on VQA-Fed.}
\label{tab:interpretability}
\begin{tabular}{lcc}
\hline
\textbf{Method} & \textbf{EC (\%)} & \textbf{FS (\%)} \\
\hline
FedAvg & 67.2 & 61.8 \\
FedPercept & 72.5 & 66.9 \\
\textbf{FedMM-X (Ours)} & \textbf{81.3} & \textbf{76.4} \\
\hline
\end{tabular}
\end{table}

\subsection{Trust Calibration and Robustness}

\begin{figure}[ht]
    \centering
    \includegraphics[width=0.9\linewidth]{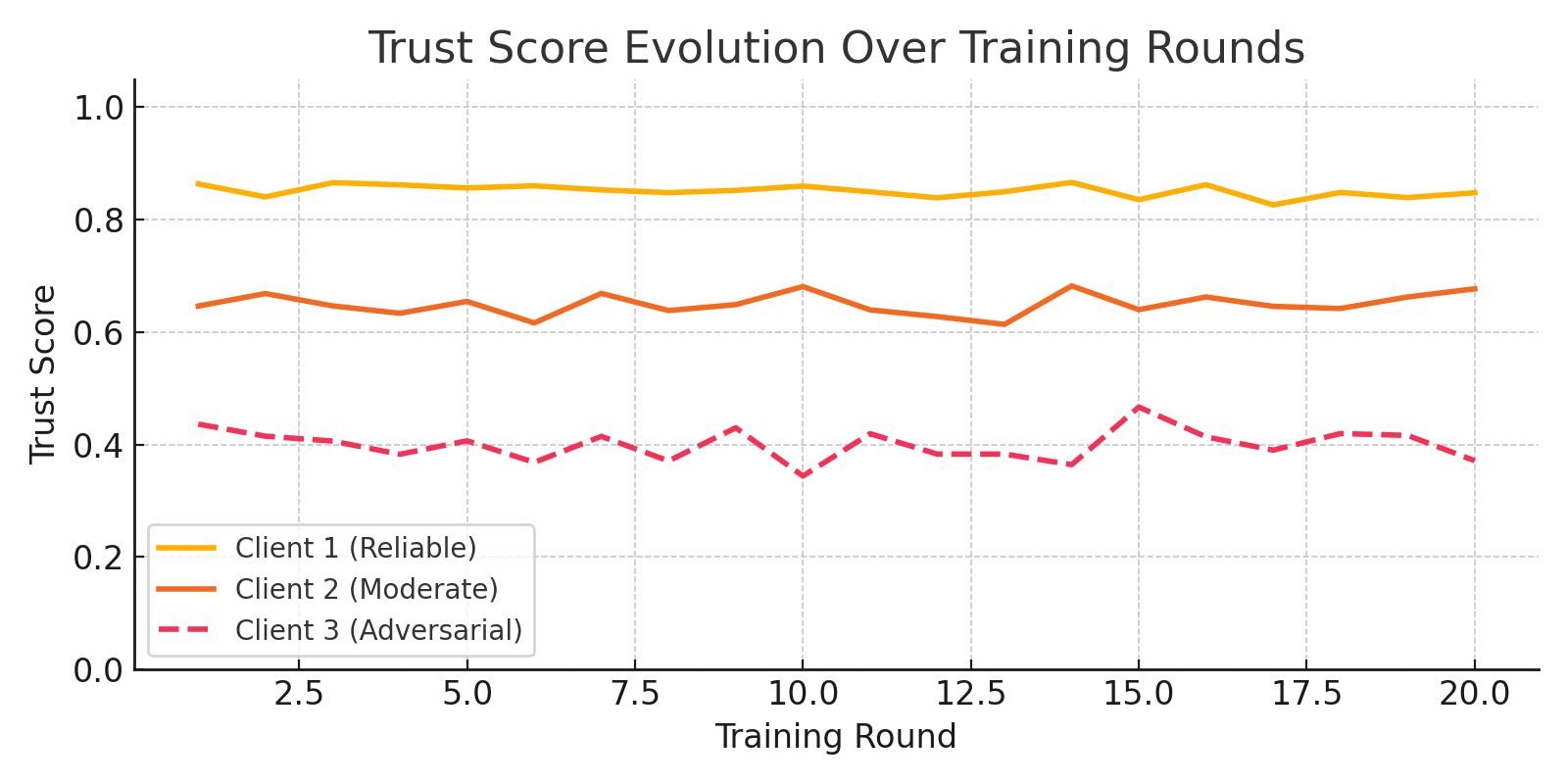}
    \caption{Trust score evolution across rounds. FedMM-X dynamically reduces the influence of noisy clients.}
    \label{fig:trust_curve}
\end{figure}

We simulate adversarial clients and missing modalities to evaluate trust-aware aggregation. FedMM-X introduces a dynamic trust score $T_k$ based on explanation coherence, model uncertainty, and prediction confidence. In presence of 20\% adversarial clients, FedMM-X maintains 91.2\% of original accuracy, compared to 78.4\% for FedAvg and 83.0\% for FedPercept. Figure~\ref{fig:trust_curve} shows how trust scores evolve over training rounds.

\subsection{Ablation Studies}

We conduct ablation studies to isolate the contribution of each component:
\begin{itemize}
    \item Removing the interpretability module reduces explanation consistency by 18.2\%.
    \item Disabling trust-aware aggregation reduces adversarial robustness by 11.6\%.
    \item Excluding cross-modal regularization reduces overall accuracy by 4.3\%.
\end{itemize}

\vspace{1mm}
\noindent FedMM-X effectively balances performance, interpretability, and trust in federated multi-modal learning, making it well-suited for deployment in dynamic and privacy-preserving environments.

\section{Conclusion and Future Work}

In this work, we presented \textbf{FedMM-X}, a novel federated multi-modal learning framework designed to ensure trustworthiness, interpretability, and robustness in Real-world AI environments. Our system uniquely integrates explainability techniques, cross-modal consistency learning, and a dynamic trust calibration mechanism to address critical challenges in decentralized learning over heterogeneous and modality-incomplete data.

We demonstrated that FedMM-X not only improves predictive performance across vision-language and audio-visual tasks but also enables interpretable local decision-making via Neural Additive Models (NAMs) and attention-based explanations. Furthermore, our trust-aware aggregation strategy effectively mitigates the negative impact of unreliable or adversarial clients an essential requirement for real-world federated deployments. The consistent improvement across metrics such as VQA accuracy, captioning scores (BLEU/CIDEr), and explanation consistency validates our claim that trust and transparency need not come at the expense of model performance.

\textbf{Key Contributions:}
\begin{itemize}
    \item We formulate the problem of \textit{trustworthy federated multi-modal learning} and propose a principled solution via FedMM-X.
    \item We incorporate interpretable reasoning models at the client level to provide feature- and modality-level explanations.
    \item We introduce a trust calibration module that enables resilient global aggregation based on explanation consistency and predictive confidence.
    \item We benchmark on three federated multi-modal datasets and provide detailed analysis through ablations and robustness evaluations.
\end{itemize}

\textbf{Limitations and Future Directions:}  
Although FedMM-X provides a promising step toward trustworthy Real-world multi-modal intelligence, several challenges remain. First, our current implementation supports only up to three modalities and assumes fixed communication rounds. Future work could explore adaptive communication schedules and asynchronous updates, which are more reflective of real-world deployments.

Second, while we utilize NAMs and attention maps for interpretability, integrating causal reasoning or concept-based explanations (e.g., TCAV) could offer deeper insights into model behavior. Moreover, extending trust calibration to account for fairness across demographic or geographic client clusters would improve the social responsibility of such systems.

Finally, we plan to apply FedMM-X to more complex multi-modal settings involving video, sensor, and temporal data, particularly in domains like healthcare, smart cities, and robotics. Real-world generalization in the presence of continuous domain shifts remains a critical direction for future research.

\vspace{1mm}
\noindent In summary, FedMM-X paves the way for next-generation federated learning systems that are not only accurate and private,

\subsubsection*{Acknowledgments}
We thank all the reviewers and mentors who provided valuable insights into our work.

\bibliography{iclr2025_conference}
\bibliographystyle{iclr2025_conference}

\end{document}